\documentclass[conference]{IEEEtran}
\IEEEoverridecommandlockouts
\usepackage{cite}
\usepackage{amsmath,amssymb,amsfonts}
\usepackage{algorithmic}
\usepackage{graphicx}
\usepackage{subcaption}
\usepackage{textcomp}
\usepackage{xcolor}
\usepackage{xspace}
\usepackage{siunitx}
\usepackage{balance}
\usepackage{enumitem}
\usepackage{caption}
\usepackage{multirow}
\usepackage{booktabs}
\usepackage{hyperref}
\usepackage{cleveref}
\usepackage{pifont} 

\usepackage{tikz} 

\def\BibTeX{{\rm B\kern-.05em{\sc i\kern-.025em b}\kern-.08em
    T\kern-.1667em\lower.7ex\hbox{E}\kern-.125emX}}

\usepackage{eso-pic}
\usepackage{url}
\AddToShipoutPictureBG*{
  \AtPageUpperLeft{%
    \put(0,-40){\raisebox{15pt}{\makebox[\paperwidth]{\begin{minipage}{21cm}\centering
      \textcolor{gray}{This article has been accepted for publication in the proceedings of the \\2024 IEEE International Instrumentation and Measurement Technology Conference (I2MTC).} 
    \end{minipage}}}}%
  }
  \AtPageLowerLeft{%
    \raisebox{25pt}{\makebox[\paperwidth]{\begin{minipage}{21cm}\centering
      \textcolor{gray}{ \copyright 2024 IEEE.  Personal use of this material is permitted.  Permission from IEEE must be obtained for all other uses, in any current or future media, including reprinting/republishing this material for advertising or promotional purposes, creating new collective works, for resale or redistribution to servers or lists, or reuse of any copyrighted component of this work in other works.
      }
    \end{minipage}}}%
  }
}

\begin{document}

\title{Towards Real-Time Fast Unmanned Aerial Vehicle Detection Using Dynamic Vision Sensors\\
}

\author{\IEEEauthorblockN{Jakub Mandula, Jonas Kühne, Luca Pascarella, Michele Magno}
\IEEEauthorblockA{\textit{Dept. of Information Technology and Electrical Engineering, ETH Zürich, Zürich, Switzerland} \\
jakub.mandula@pbl.ee.ethz.ch, kuehnej@ethz.ch, luca.pascarella@pbl.ee.ethz.ch, michele.magno@pbl.ee.ethz.ch }
}

\newcommand{\ie}{\emph{i.e.,}\xspace}
\newcommand{\eg}{\emph{e.g.,}\xspace}
\newcommand{\vs}{\emph{vs.}\xspace}
\newcommand{\etc}{etc.\xspace}
\newcommand{\etal}{\emph{et~al.}\xspace}
\newcommand{\todo}[1]{\textcolor{red}{TODO: #1}}
\newcommand{\rev}[1]{#1}

\newcommand*\mycirc[1]{%
  \begin{tikzpicture}[baseline=(C.base)]
    \protect\node[draw,circle,inner sep=1pt](C) {#1};
  \end{tikzpicture}}

\newcommand{\tool}{{\textsc{\textcolor{black}{F-UAV-D}}}\xspace}
\newcommand{\toolFull}{{\textcolor{black}{Fast Unmanned Aerial Vehicle Detector}}\xspace}

\newcommand{\update}[1]{\textcolor{blue}{#1}}

\maketitle

\begin{abstract}
Unmanned Aerial Vehicles (UAVs) are gaining popularity in civil and military applications. However, uncontrolled access to restricted areas threatens privacy and security. Thus, prevention and detection of UAVs are pivotal to guarantee confidentiality and safety. Although active scanning, mainly based on radars, is one of the most accurate technologies, it can be expensive and less versatile than passive inspections, \eg object recognition. Dynamic vision sensors (DVS) are bio-inspired event-based vision models that leverage timestamped pixel-level brightness changes in fast-moving scenes that adapt well to low-latency object detection. This paper presents \tool (\toolFull), an embedded system that enables fast-moving drone detection. In particular, we propose a setup to exploit DVS as an alternative to RGB cameras in a real-time and low-power configuration. Our approach leverages the high-dynamic range (HDR) and background suppression of DVS and, when trained with various fast-moving drones, outperforms RGB input in suboptimal ambient conditions such as low illumination and fast-moving scenes. Our results show that \tool can (i) detect drones by using less than \SI{<15}{\watt} on average and (ii) perform real-time inference (\ie \SI{<50}{ms}) by leveraging the CPU and GPU nodes of our edge computer.
\end{abstract}


\DeclareSIUnit\pixel{px}

\section{Introduction} \label{sec:intro}

Unmanned Aerial Vehicles (UAVs), due to their versatility, high mobility, and low cost, are gaining popularity in various scenarios spanning from leisure to rescue missions or keeping out intruders~\cite{mohsan2022towards}. UAVs show their superiority in civil applications to meet strict targets, such as reducing the CO2 footprint of human-centered activities (\eg parcel delivery~\cite{li2022application} or precision agriculture~\cite{fahey2021integration}), or even in military applications such as military surveillance~\cite{utsav2021iot}.
Nonetheless, illegal trespassing of no-flying zones, such as border encroachment or illegal overflights over protected areas, is a massive predicament against privacy and security~\cite{bonetto2015privacy,shi2018anti}.

The literature proposes various drone detection techniques, mainly falling into two categories based on active and passive detection~\cite{taha2019machine}. Active detection leverages computational analysis of radar signal reflection~\cite{musa2019review}. Grounded on well-known technology, radars guarantee high accuracy in object localization, especially when employing high-frequency waves (\eg mmWave~\cite{fu2021deep,morris2021detection}). Nevertheless, active scanning is typically energy-demanding, and an accurate resolution requires costly equipment. A promising, passive, and cost-effective solution is based on acoustic detection. In this case, an array of microphones can be deployed to isolate and recognize the noise of propellers~\cite{shi2020acoustic}. On the other hand, it is sensitive to ambient noise, especially in loud areas, such as airports~\cite{khan2022detection}. A second passive UAV detection solution exploits visual scanning. A video camera produces a stream of frames, and a computer algorithm continuously searches for unauthorized flying objects by emulating a human inspector.

A pioneer work in video-based drone detection has been introduced by Rozantsev \etal~\cite{rozantsev2016detecting}. The authors propose two methods for detecting UAVs by using a single camera and a Convolutional Neural Network (CNN) model. By exploiting the potential of deep learning, and in particular, the well-known object detection algorithm YOLO~\cite{terven2023comprehensive}, other researchers such as Alsanad \etal~\cite{alsanad2022yolo} and Singha \etal~\cite{singha2021automated} tackled the same problem, exceeding 90\% of precision and recall given a static background. More sophisticated CNNs and data examples can help researchers outperform previous results. However, camera-based object recognition still suffers under suboptimal testing conditions~\cite{pascarella2023grayscale}.

In contrast to RGB cameras, event-based vision is a set of biologically inspired vision algorithms driven by continuous scene changes.
The dynamic vision sensor (DVS)~\cite{serrano2013128,gallego2018} implements high-speed event generation in hardware acting as a high-pass filter that filters static and consequently redundant information out from a scene. 
A DVS generates an independent response to brightness changes for each pixel containing timing, intensity, and matrix-level position.
Due to the intrinsic characteristics,
event-based vision algorithms perfectly fulfill tasks with dynamic and fast-moving scenes~\cite{gallego2020event,sandamirskaya2022neuromorphic}.

\rev{This paper presents \tool (\toolFull) as an approach to enable autonomous detection of flying drones in challenging static scenarios using an embedded system to minimize latency and power consumption. }
\rev{The primary objective is to enable the deployability at the edge when detecting fast-moving UAVs ($>$\SI{1}{\kilo\pixel\per\second}) using event-based frames. We also compared various batch sizes to determine the optimal latency and power consumption balance. In pursuing this objective, we created a ground truth dataset, which we made publicly available~\cite{FDDPublicDroneDataset}.}

\textbf{Research contribution.} \tool is the next step to design a portable and low-power UAV detector.
The contributions are:

\begin{itemize}

\item Design and development of an embedded system for energy-efficient and low latency UAV detection;
\item  Dataset collection and ground truth creation for machine learning training and evaluation;
\item An empirical evaluation of the power consumption at different batch sizes of YOLOv8 \rev{at the edge}.

\end{itemize}



\section{Related Work} \label{sec:related}

We focus our discussion on (i) event-based sensors, (ii) available datasets, and (iii) deep-learning approaches for DVS. Due to space limitations, we omit the discussion of the many applications of DVS that are less relevant to this study, pointing the reader to an inclusive survey by Gallego \etal~\cite{gallego2020event} and a review by Sandamirskaya \etal~\cite{sandamirskaya2022neuromorphic}.


The RGB camera---that takes its name from red, green, and blue color schema---is designed to return brightness information in the form of a matrix of pixels at a constant interval known as frame rate (\eg 30 fps) in the full visible color spectrum.
\rev{Event-based sensors, such as DVS~\cite{lichtsteiner2008,lichtsteiner200564x64}, output a stream of events in response to brightness changes.
Each event carries information about one pixel at a time (\ie the one affected by the brightness variation in the scene) in addition to a timestamp. More precisely, an event indicates the planar pixel coordinates, a timestamp, and the polarity of the brightness change (positive and negative).}
Thus, the throughput is variable (up to 450\,M events-per-second~\cite{iniVationHome}) and proportional to the number of captured events. 
DVS presents advantages regarding \textit{high dynamic range} (HDR), up to 140dB, that allows for capturing bright stimuli even in dark environments~\cite{brandli2014real,gallego2018}, \textit{low-power} impact $<$10mW due to the absence of redundant information~\cite{amir2017low}, and \textit{low-latency} response, $<$10µs, because of each pixel working independently~\cite{brandli2014real,everding2018low}. 
These characteristics make DVS outperform RGB cameras in challenging scenarios. Indeed, researchers propose several studies on event-based computer vision. For example, Orchard \etal~\cite{2015orchardConvertingStaticImage} presented a method for converting existing images into event-based datasets using an actuated pan-tilt camera platform. Mueggler \etal~\cite{2017muegglerEventCameraDatasetSimulator} and de Tournemir \etal~\cite{2020detournemireLargeScaleEventbased} created two real datasets using two different DVS sensors.
Zhang \etal~\cite{2021zhangObjectTrackingJointly} conducted a study on object tracking by using a dataset generated with a DAVIS368 camera. Nonetheless, none of these studies focus on UAVs. One of the first studies framing UAVs has been introduced by Wang \etal~\cite{2021wangVisEventReliableObject}. The study extensively compares different machine-learning models, highlighting some limitations of object tracking when used with event-based images.

\begin{table}[h]
    \begin{tabular}{lcccc}
        \toprule
        \textbf{Dataset/Year}                                               & \textbf{Resolution}  & \textbf{UAV/Real} & \textbf{DVS} & \textbf{Hours}  \\
        \midrule
        N-MNIST~\cite{2015orchardConvertingStaticImage} `15                 & 304x240              & \ding{56}      \ding{56}     & ATIS      & $<$1   \\
        EED~\cite{2017muegglerEventCameraDatasetSimulator} `18              & 240x180              & \ding{56}      \ding{52}     & DAVIS     & $<$1   \\
        Automotive~\cite{2020detournemireLargeScaleEventbased} `20          & 1280x720             & \ding{56}      \ding{52}     & Prophesee & $<$40   \\
        FE108~\cite{2021zhangObjectTrackingJointly} `21                     & 346x260              & \ding{56}      \ding{52}     & DAVIS368  & 1.5   \\
        VisEvent~\cite{2021wangVisEventReliableObject} `21                  & 346x260              & \ding{52}      \ding{52}     & DAVIS368  & $<$5   \\
        Our dataset~\cite{FDDPublicDroneDataset} `23                        & 1280x720             & \ding{52}      \ding{52}     & Prophesee &   0.5  \\
        \bottomrule
        \end{tabular}
    \centering
    \caption{Comparison of various event-based datasets.}
    \label{tab:datasets}
\end{table}

\Cref{tab:datasets} summarizes the characteristics of DVS-driven and UAV-dedicated studies. In particular, for our study, we continued in the direction of using deep-learning models to detect fast-moving drones by experimenting with state-of-the-art YOLO.
Finally, although the literature proposes event-based datasets of diverse subjects, including UAVs, they do not fit our study goal (\ie synthetic, low resolution, or do not focus on UAVs). Thus, we created a publicly available dataset~\cite{FDDPublicDroneDataset} suited for our study that includes 27\,min of video of drones in a closed and open environment.




\section{System Overview} \label{sec:overview}

\begin{figure}[t]
    \centering
    \includegraphics[width=1\columnwidth]{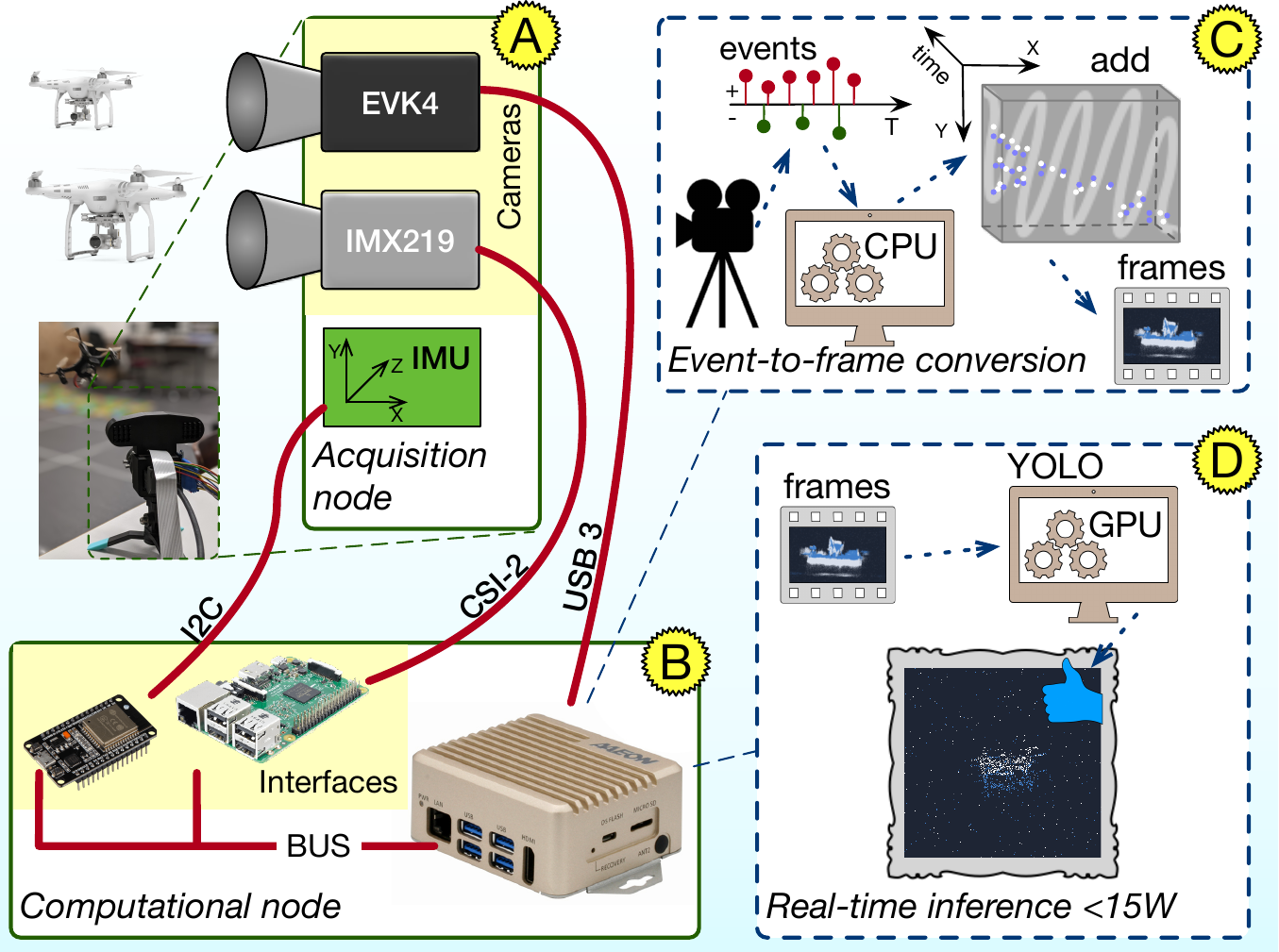}
    \caption{Overview of the system. The solid boxes \mycirc{A} and \mycirc{B} show the hardware components and the dashed boxes \mycirc{C} and \mycirc{D} illustrate the software parts.}
    \label{fig:system}
\end{figure}

\rev{This paper presents \tool as a frame-based approach for conventional deep learning algorithms that are trained to detect fast-moving UAVs at the edge in real-time, using event-to-frame inputs.} The following sections report the hardware and software setup depicted in \Cref{fig:system}.

\subsection{Hardware} \label{sec:overview:hardware}

\begin{figure}[t]
    \centering

    \begin{subfigure}[b]{0.15\textwidth}
        \includegraphics[width=\textwidth]{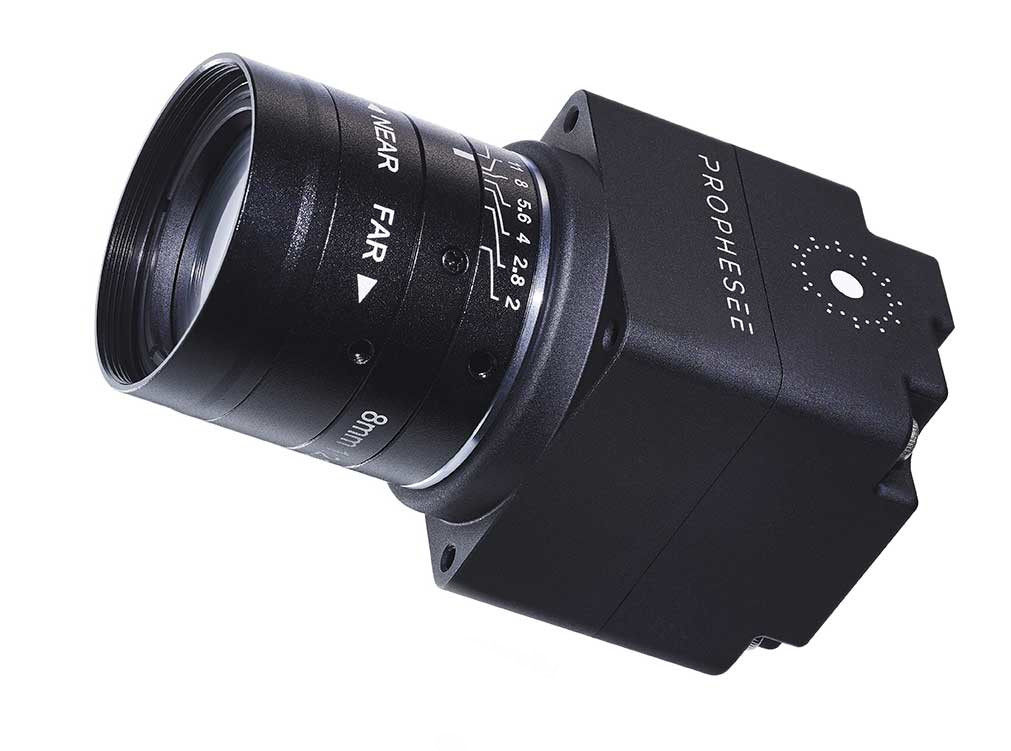}
        \caption{EVK4}
        \label{fig:evk4}
    \end{subfigure}
    \begin{subfigure}[b]{0.15\textwidth}
        \includegraphics[width=\textwidth]{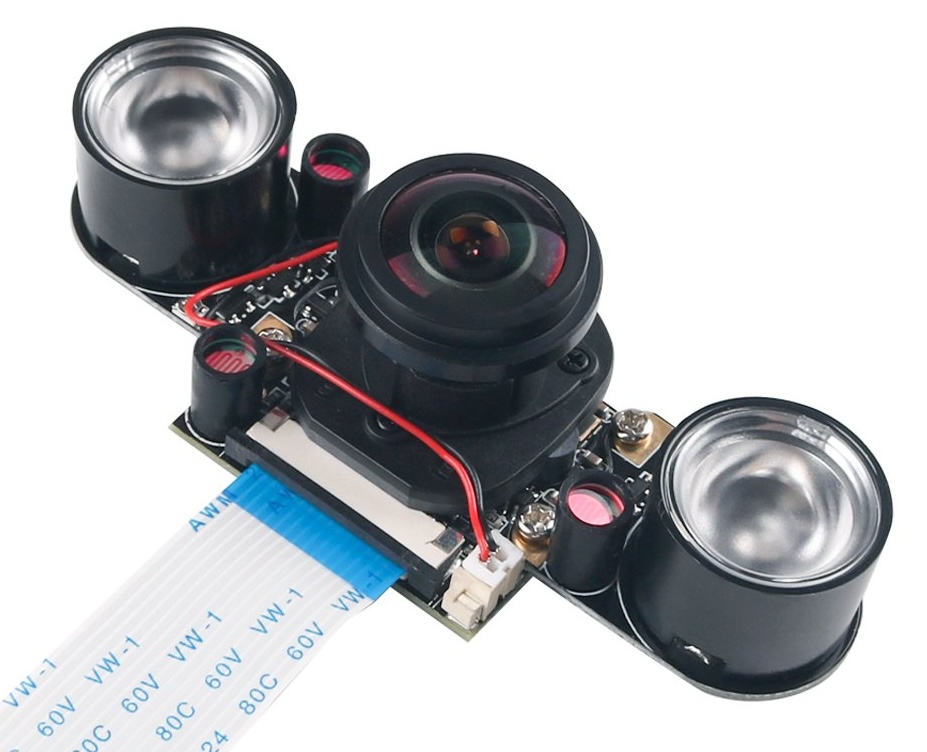}
        \caption{RPi Camera}
        \label{fig:imx219}
    \end{subfigure}
    \begin{subfigure}[b]{0.15\textwidth}
        \includegraphics[width=\textwidth]{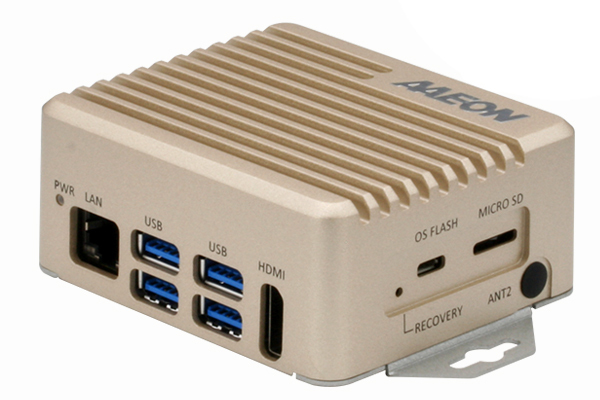}
        \caption{BOXER~8251}
        \label{fig:boxer8251}
    \end{subfigure}
    \caption{Selected image sensors and computational unit.}
    \label{fig:hardware_setup}
    \vspace{-0.4cm}
\end{figure}

\rev{\tool is built on top of a commercial DVS camera, a CMOS RGB array sensor, and a System-on-Module (SoM) low-power computation node. These elements compose the logical blocks \mycirc{A} and \mycirc{B} in \Cref{fig:system}. \Cref{fig:evk4} shows the ultra high-speed, HDR event-based vision evaluation kit EVK4~\cite{2023EventbasedVisionEvaluation} selected for this study, which is built on top of Sony's new IMX636 sensor. The camera was mounted with a \SI{8}{\milli\meter} lens.}
\rev{\Cref{fig:imx219} shows a Raspberry Pi camera based on Sony's IMX219 RGB sensor with two additional infrared LEDs for night vision (not used in our experiment). The camera was mounted with a \SI{6}{\milli\meter} lens and generates $1280\times720$ pixel frames at 30\,fps, with a back-illuminated CMOS array with an RGB primary color mosaic filter. The IMX219 offers a MIPI CSI-2 flex cable interface.} 
Finally, \Cref{fig:boxer8251} reveals a low-power edge computer capable of speeding up highly parallelizable computations such as deep learning algorithms. The AAEON BOXER-8251AI~\cite{2023AAEONBOXER8251AIAI} is a passive-cooled case that hosts an Nvidia Jetson Xavier NX SoM module. This ARM-based SoC integrates a 64-bit 6-core heterogeneous CPU together with a 384-core Volta architecture GPU, 48 Tensor Cores, and a shared 8GB of LPDDR4x RAM. This allows the relatively small module to reach a performance of about 21 TOPS using INT8 arithmetic at only \SI{15}{\watt}. A 12V socket powers the BOXER-8251AI and makes it field-deployable.
\rev{In addition, an Inertial Measurement Unit (IMU) with 9 degrees of freedom, composed of LSM303DLHC (\ie 3-axis accelerometer and 3-axis magnetometer) and L3GD20 (\ie 3-axis Gyroscope) has also been added to the acquisition node to capture movements with up to 220Hz and enable a post-acquisition compensation of ego-motion (\eg vibrations).}

\subsection{Software} \label{sec:overview:software}

From the software perspective, \tool is based on a combination of off-the-shelf and custom tools, including a real-time object detection algorithm and a routine for event-to-frame conversions. In \Cref{fig:system}, \mycirc{C} depicts the event-to-frame conversion (see \Cref{sec:accumulation}) happening in real-time on the CPU (\ie running at 1.9 GHz) of the BOXER 8251. \mycirc{D} summarizes the logical flow in the object detection algorithm. To optimize the real-time performance as well as the energy consumption, the inference algorithm runs on the GPU (\ie 1.1 GHz) that is extremely specialized to execute highly parallelizable tasks, such as the case of CNN-based inference.


\section{Methodology} \label{sec:methodology}

The \emph{goal} of this paper is to empirically assess the performance of the proposed event-based computer vision algorithm in real-time, running on a low-end computing system, in detecting fast-moving UAVs. The \emph{context} is represented by the dataset collected using the hardware described in \Cref{sec:overview:hardware}. The study aims to answer our research question:

\begin{description}

\item[\textbf{RQ:}]\textit{To what extent do different batch sizes influence power consumption and inference latency in embedded systems?} With RQ, we compare different batch sizes to assess the best compromise between latency and power consumption when running inference on an embedded system.
\end{description}

\subsection{Dataset Labeling} \label{sec:dataset}

Creating a large ground truth dataset is expensive and tedious due to recurring manual operations. Nevertheless, it becomes a \emph{dictate} when novel sensors are exploited to train machine learning algorithms~\cite{paullada2021data}. In addition, labeling event-based frames is more challenging due to the artificial blur generated by event accumulation. However, by having RGB and DVS inputs synchronized, the annotator can always inspect one of the two inputs to have visual confirmation (\eg \Cref{fig:eventComparison}).
To speed up the manual activity, we leveraged Label Studio~\cite{2020tkachenkoLabelStudioData}, an open-source general-purpose labeling tool designed to support multi-user manual labeling of various inputs \eg audio, video, text \etc
Label Studio supports the keyframe interpolation. In other words, the highlighted subject (\eg a drone) does not need to be labeled in every single frame, but a linear interpolation is performed across frames. 

\subsection{Event-to-Frame Conversion} \label{sec:accumulation}

\begin{figure}[t]
    \centering
    \begin{subfigure}[b]{0.23\textwidth}
        \includegraphics[width=\textwidth]{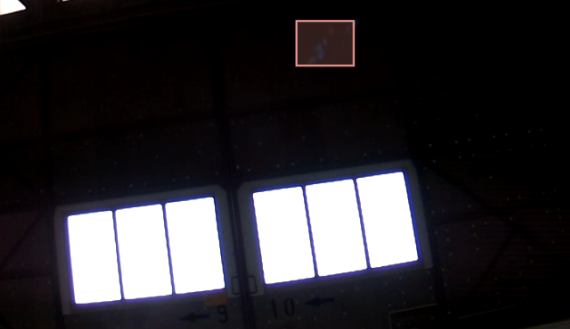}
        \caption{RGB frame.}
        \label{fig:rgb-frame}
    \end{subfigure}
    \begin{subfigure}[b]{0.23\textwidth}
        \includegraphics[width=\textwidth]{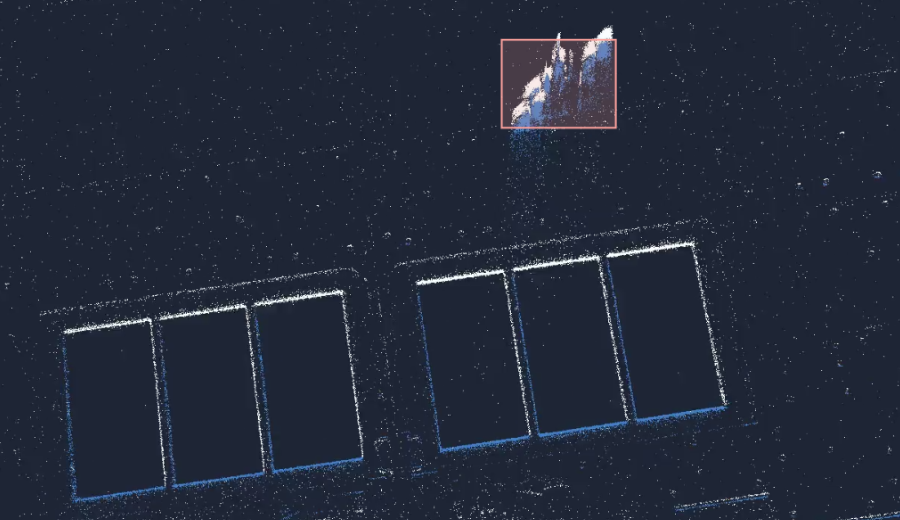}
        \caption{Event-to-frame.}
        \label{fig:event-to-frame}
    \end{subfigure}
    \caption{An example of RGB and event-to-frame overlapping.}
    \label{fig:eventComparison}
\end{figure}

Conventional computer vision algorithms support inputs in the form of frames (\eg a 2D matrix of pixel intensities). This input format corresponds to the standard RGB camera output. On the contrary, a DVS generates a stream of events in response to a brightness change. Thus, practitioners rely on a preprocessing step to adapt event-based output to conventional computer vision algorithms, where events are accumulated to generate artificial frames~\cite{binas2017ddd17}. The result is a stream (at a constant rate) of a 2D matrix of pixels' intensity changes. Specifically, for our approach, we accumulated in a 2D matrix $m$ the events $e_k=(x_k,y_k,t_k,p_k)$ corresponding to all brightness changes happening in the predefined time interval $T$. It is worth noticing that to prevent pixel-wise information loss due to the accumulation of both positive and negative brightness changes in the same time window, we accumulated positive and negative events into two different 2D matrices. 
In other words, for every accumulation interval $T$, we generated a 2-color frame, white for positive events and blue for negative ones. \Cref{fig:eventComparison} shows a frame where a drone is focused by a conventional RGB camera (left) and by a DVS (right). In particular, \Cref{fig:event-to-frame} highlights the artificial frame generated by accumulating events over an interval $T$. \rev{It is worth mentioning that although there are alternative frame-to-event conversion techniques, we chose the above one, as it is straightforward to implement, while still retaining as much information as possible by using two color channels.}

\subsection{Camera Synchronization and Calibration} \label{sec:sync_and_calibration}

\begin{figure}[t]
    \centering
    \begin{subfigure}[b]{0.23\textwidth}
        \includegraphics[width=\textwidth]{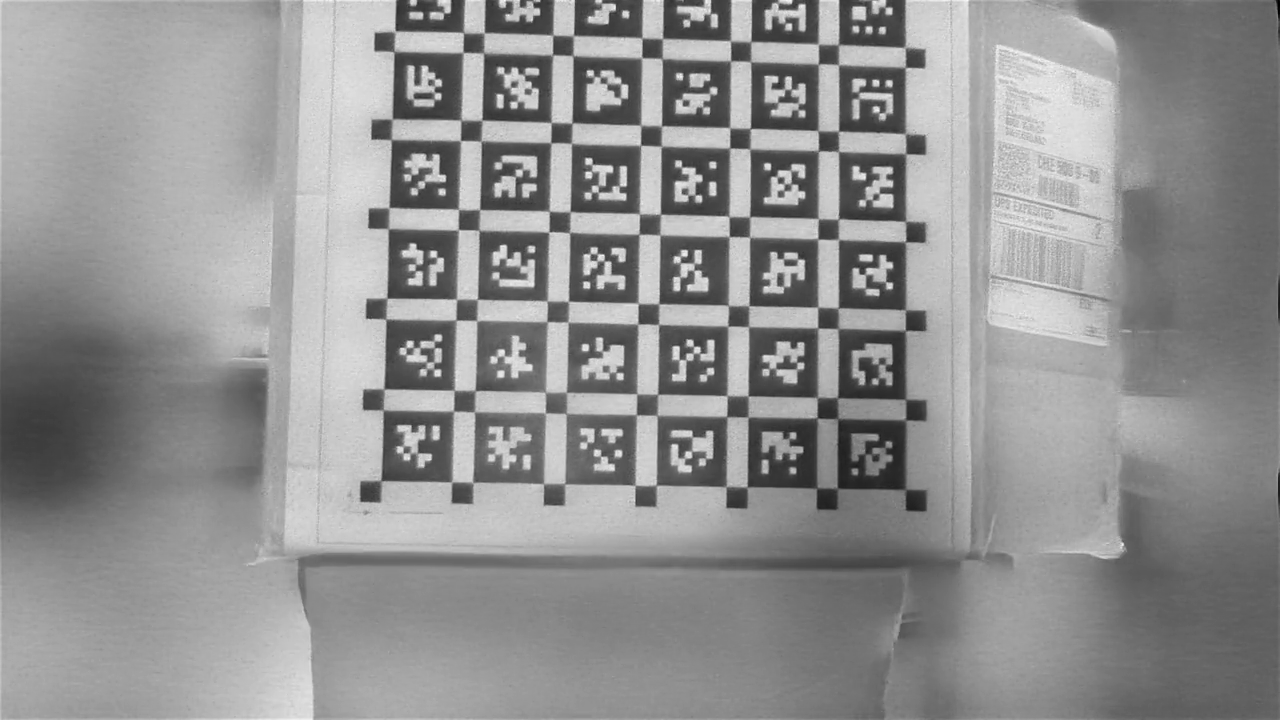}
        \caption{RGB frame.}
        \label{fig:aprilTagRgb}
    \end{subfigure}
    \begin{subfigure}[b]{0.23\textwidth}
        \includegraphics[width=\textwidth]{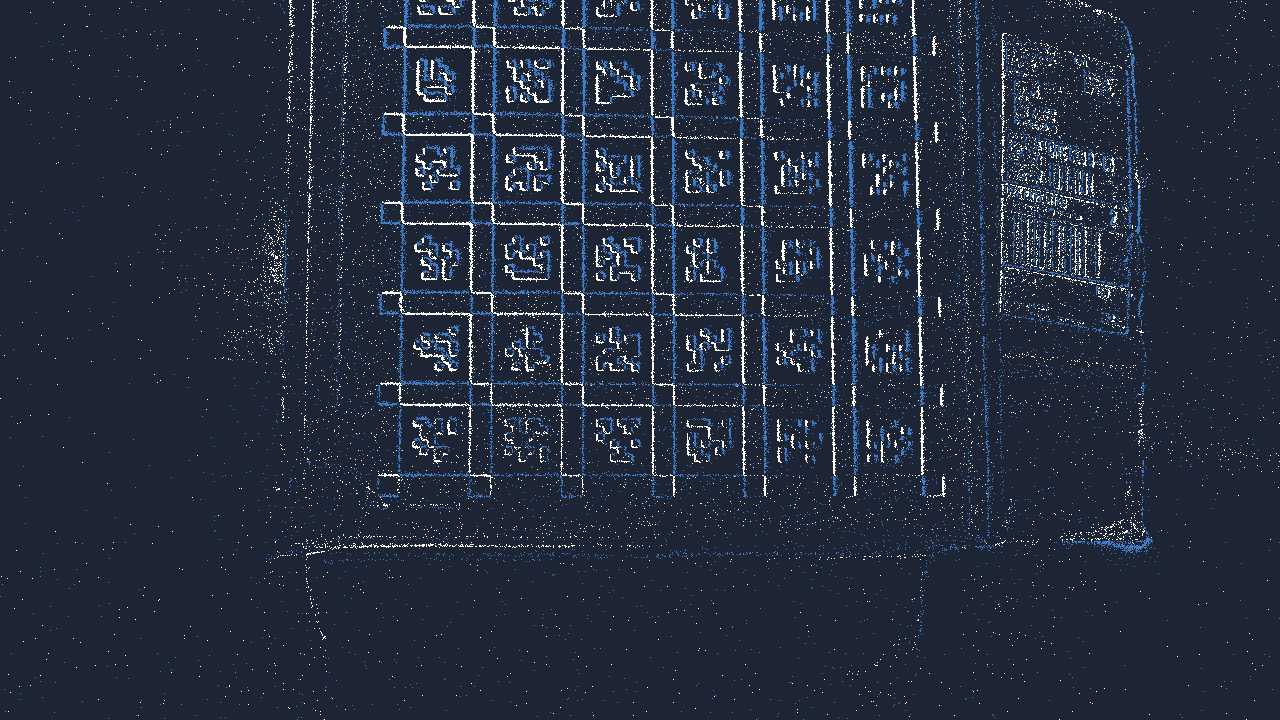}
        \caption{Event-to-frame.}
        \label{fig:aprilTagDvs}
    \end{subfigure}
    \caption{AprilTag grid in RGB and event-to-frame view.}
    \label{fig:aprilTags}
\end{figure}

We applied an image transformation to align the RGB and event-to-frame viewpoints. This (i) allows the manual annotator to tag drones in a format (\eg RGB) and transfer the coordinates in the other (\eg DVS) and (ii) enables an automatic labeling process by running a detector in RGB space and transferring the labels to event space. The transformation is necessary due to the different optics in the IMX219 and EVK4 cameras.
\rev{It is worth mentioning that the IMX219 is a rolling shutter camera that introduces artifacts in fast-moving objects. Nonetheless, these rolling shutter artifacts do not represent an issue for our study because we use a different input (\ie DVS) for drone detection. At the same time, IMX219 still represents an inexpensive, plug-and-play, and off-the-shelf camera that is convenient for generating reference labels.}
However, the first challenge before applying the mathematical transformation model is having perfect time synchronization.
We relied on Zero-mean Normalized Cross-Correlation (ZNCC), a popular and robust correlation/distortion function often used for template matching between two images~\cite{di2005zncc}. ZNCC provides a similarity coefficient between two sample images; the higher the coefficient, the closer the two images are.
We compute the temporal offset between the two sequences by maximizing the ZNCC value of the two streams. However, it is worth mentioning that the DVS camera has a synchronization port that allows the injection of artificial events into the event stream. In other words, the injected sequences generate time-stamped events that simplify the post-acquisition time alignment.
To apply the pinhole camera model~\cite{juarez2020distorted} and have an accurate mapping between the RGB and DVS frames, we needed to retrieve (i) the camera's intrinsic parameters $K_1$ and $K_2$, (ii) the mutual camera-to-camera extrinsic parameters $\,_1T_2 = \,_1[R\; t]_2$, and (iii) the lens distortion models $d_1$ and $d_2$.
To find the camera's parameters, we relied on the Kalibr calibration toolbox~\cite{2016rehderExtendingKalibrCalibrating}, which uses the synchronized videos of a moving AprilTag grid (shown in \Cref{fig:aprilTags}) to measure the intrinsic and extrinsic parameters, as well as the lens distortion. However, Kalibr does not support a stream of events out of the box; therefore, we applied the technique adopted by Muglikar \etal~\cite{2021muglikarHowCalibrateYour}. In summary, we (i) generated the frames by accumulating events and (ii) applied the Kalibr technique.

\subsection{Data Collection} \label{sec:data_collection}

\rev{Even though the growing popularity of DVS in research has led to the creation of publicly available datasets (\eg \Cref{tab:datasets}), the literature lacks datasets focusing on UAVs.}
For this reason, we relied on the setup described in \Cref{sec:overview:hardware} to create a publicly available dataset of different drones~\cite{FDDPublicDroneDataset}.
To maximize the storage efficiency and prevent bandwidth limitations, we used \textsc{zstd}~\cite{collet2018zstandard} for an on-the-fly compression while using Linux BTRFS~\cite{rodeh2013btrfs} as a storage file system. For efficient labeling of the newly recorded dataset, we used the YOLO detector~\cite{redmon2016you} on the conventional camera frames. Using the camera-to-camera calibration discussed in \Cref{sec:sync_and_calibration} and assuming that the objects are very far away (\ie assuming zero disparity), the labels obtained in the conventional frame can be transformed to the event-camera frame to be used as a prior.
We assessed the performance of \tool using the mAP (mean Average Precision) and measured the inference time and power.
On top of our quantitative analysis, we also performed a qualitative analysis to better understand the strengths and weaknesses of \tool. We manually inspected a set of ``inaccurate predictions'' \eg predictions in which the model detects a drone but the position and dimension of the bounding box are inaccurately predicted.

\section{Results} \label{sec:results}

To answer our RQ we run \tool using the dataset described in \Cref{sec:data_collection}. Thus, we create a ground truth of manually labeled flying drones (\ie \Cref{sec:dataset}). Successively, we train \tool to evaluate the characteristics of the proposed event-based deep learning object detection algorithm in terms of power consumption and inference time.

\textbf{Event-to-frame conversion.} Conventional computer vision algorithms, such as YOLO~\cite{terven2023comprehensive}, are designed to work with frame-based inputs that typically match the output of RGB cameras. \rev{The model expects as input a 3D matrix (\eg a 2D matrix with three color channels) of pixels brightnesses, \ie $1280\times720$ for \tool. However, in event-based approaches, the output of the DVS is a stream of events.} To feed YOLO with event-based inputs, we needed to aggregate the stream of events in a frame-based manner. Specifically, we applied the straightforward accumulation process described in \Cref{sec:accumulation}. 
\rev{As the first step, we empirically evaluate different integration time intervals, spanning from 10~ms to 50~ms.}
A shorter integration time captures fewer details (because fewer occasional events appear in the frame), but the edges appear sharp. In contrast, a longer integration time translates into more details corresponding to unsharpened edges in a moving scene.
We chose a value of 33.3 ms corresponding to a frame rate of 30 fps to simplify the alignment of the event-based frames with the RGB ones (working at 30 Hz) and because it creates an acceptable blur effect.
Moreover, we experimented with C \texttt{uint8} as the data type for the accumulation step to comply with the RGB color scheme.
Finally, we defined a clipping threshold of 255; in other words, if an event for a pixel is triggered more than 255 times during the integration interval, we still report 255 as a saturated value. However, given the short interval of 33.3 ms, this never happens in practice.
This event representation allows efficient storage because it requires only $1280\times720\times2\times1 B = 1.84 MB$ of storage per single 2D event-to-frame. 

\textbf{Dataset Labeling.} The next step consists of labeling the generated frames (\ie event-to-frame \Cref{sec:accumulation}). This step is crucial to train YOLO with a ground truth dataset. For this task, we used the automatically generated label candidates within Label Studio (see \Cref{sec:dataset}), which allows a human operator to draw a bounding box around the aimed subject. 

\textbf{YOLOv5 training.} To achieve our goal, we started with the revolutionary YOLOv5 algorithm introduced in its first version by Redmon \etal back in 2015~\cite{redmon2016you} and which reached version 8 recently~\cite{terven2023comprehensive}. YOLOv5 is an up-to-date deep learning algorithm that supports real-time object detection.
YOLO stands for ``You Only Look Once'', and it is a real-time end-to-end approach for object detection referring to the fact that it achieved the detection task with a single pass of the network, as opposed to previous approaches that either
used sliding windows followed by a classifier that needed to run hundreds or thousands of times per image or the more
advanced methods that divided the task into two-steps. Also, YOLOv5 used a more straightforward output based only on regression to predict the detection outputs as opposed to Fast R-CNN~\cite{girshick2015fast} that used two separate outputs, a classification for the probabilities and a regression for the boxes coordinates.
\rev{We adapted \emph{YOLOv5-nano} with \SI{1.9}{\mega\ parameters}  to accept a two-dimensional, 2 channel matrix corresponding to our event-to-frame accumulation. Moreover, it has been trained using the default training agents and parameters from YOLOv5 and evaluated using IoU threshold of 0.5\cite{2023jocherYOLOUltralytics}.} \rev{Finally, to deploy our approach on the Jetson platform and obtain the best inference performance, we optimized the YOLO network using TensorRT Command-Line Wrapper \texttt{trtexec} ~\cite{trtexec} and quantize it.}

\textbf{Power measurements.} To measure the total power consumption of \tool during inference, we relied on the \emph{Keysight N6705C} DC Power Analyzer. We defined a current and voltage sampling of 0.5 ms. However, to filter the noisy waveform, we applied a moving average of 10 samples. 

\begin{figure}[t]
    \centering
    \includegraphics[width=.87\columnwidth]{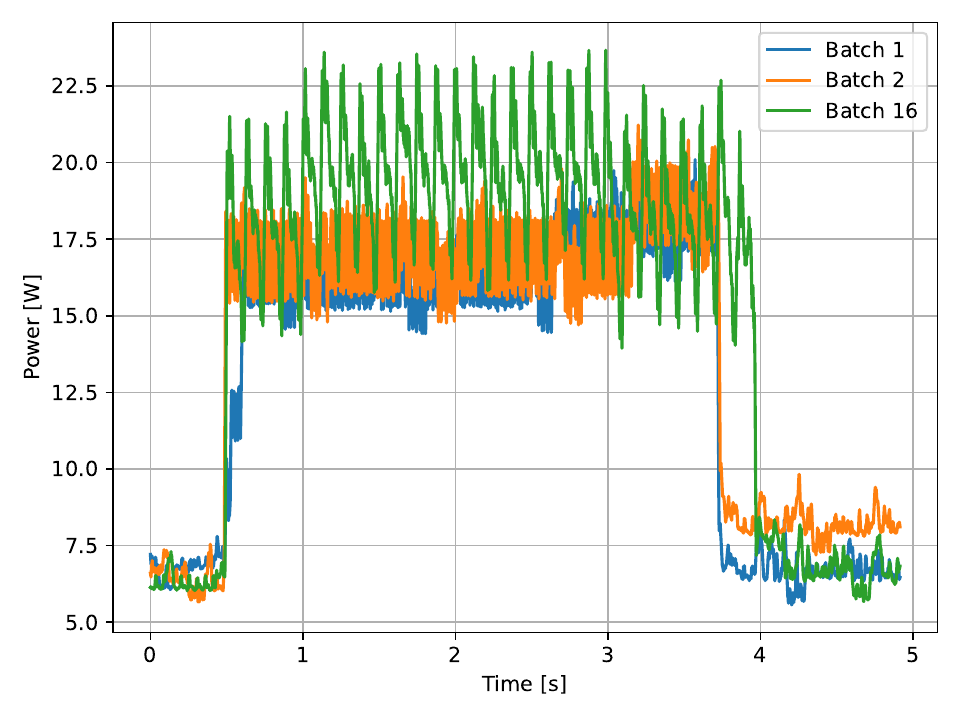}
    \caption{\rev{Power consumption during real-time inference. For batch sizes $B=1$, $B=2$, and $B=16$}}
    \label{fig:power}
\end{figure}

\begin{figure}[t]
    \centering
    \includegraphics[width=.87\columnwidth]{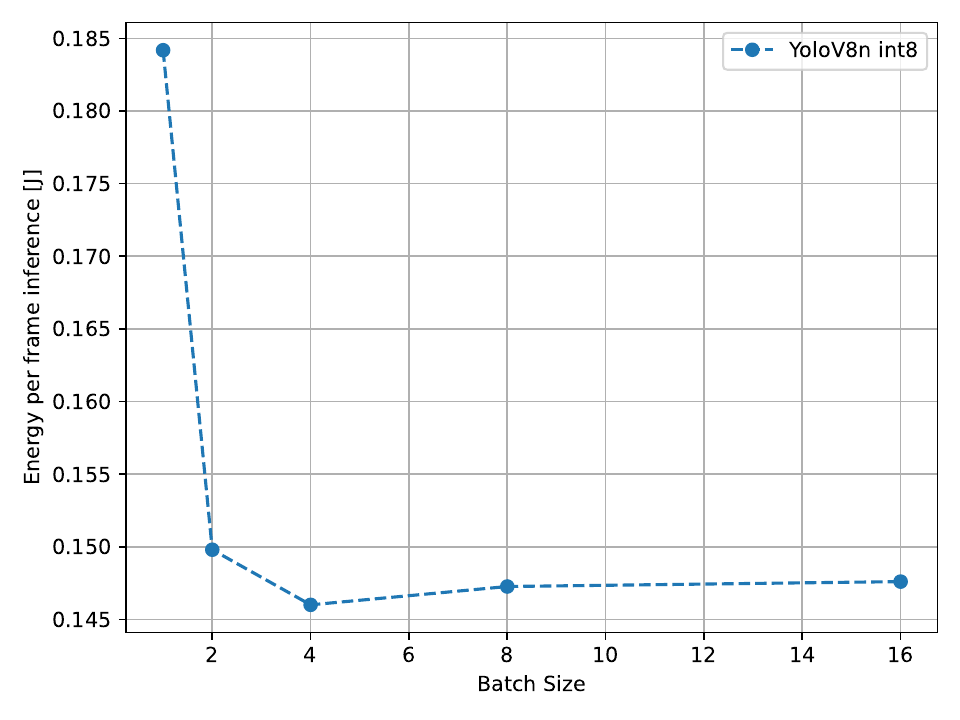}
    \caption{Energy per inference frame.}
    \label{fig:energy-per-frame}
\end{figure}

\rev{\Cref{fig:power} compares different batch sizes of the typical power waveform when running YOLOv5 on the GPU of the AAEON BOXER-8251AI. \rev{It is worth mentioning that the batched input for the adapted YOLO network is a $B\times640\times480\times2$ tensor generated from the event-to-frame conversion running in parallel on the CPU.} At first glance, \emph{Batch 16} seems to have higher current spikes, while \emph{Batch 1} sits below. However, this sawtooth waveform hides that, on average, \emph{Batch 16} is processing more data at the time. Thus, it requires less energy per frame.}
\Cref{fig:energy-per-frame} focuses on the energy demand per single inferred frame. As expected, inferring multiple frames at a time (\ie running YOLO on a batch size bigger than one) helps to reduce the average power consumption. Indeed, in contrast to what can be revealed at a first look from \Cref{fig:power}, by calculating the power absorbed per single frame as the average power divided by the batch size, the worst scenario happens exactly when the batch size contains a single frame \ie \SI{182}{\milli\joule} per frame. On the contrary, when running YOLO on a batch larger than one, the energy demand drops. The highest efficiency is reached at a batch size of four, requiring roughly \SI{146}{\milli\joule} per frame. Although a batch size greater than one allows for reduced power consumption, it comes with a drawback: a higher inference latency. \rev{Based on the benchmarks, we calculated that for the worst case (\ie Batch 16) the inference latency only increases by a factor of two. That value is still acceptable for a real-time inference because although we need to wait \SI{33.3}{\milli\second} (due to 30 fps) $\times$ 16 (number or consecutive frames) + $\sim$\SI{100}{\milli\second} (inference time) the total detection time is kept below the one-second threshold. It is worth noticing that Batch 16 is far from being the best fit in terms of power consumption. The best conditions for power and inference latency are with Batch 4, which has a frame rate delay of $\sim$\SI{133}{\milli\second} plus an additional $\sim$\SI{150}{\milli\second} for inference.}


\section{Conclusion and Future Work} \label{sec:conclusion}
In this paper, we presented \tool an embedded system consisting of hardware and software components to enable UAV detection using event-to-frame inputs. We presented a hardware setup consisting of a conventional RGB camera, a novel DVS sensor, and a low-power embedded system. We showed how existing machine learning tools like YOLO can be used to automate the labeling process for creating new event-camera datasets. \rev{In addition, we openly provided our dataset~\cite{FDDPublicDroneDataset}, consisting of 27\,min of stationary and fast-moving drone recordings (\ie 49,000 frames).} We preliminary validated the above dataset by measuring power consumption and latency.
The measured performance (\ie 0.53 mAP) shows an in-field deployability thanks to a real-time inference (\ie less than 50ms) that demands only \SI{150}{\milli\joule} per frame.

\rev{We plan to extend this work by (i) extending the dataset (\eg including more challenging conditions), (ii) conducting a deeper investigation of the proposed setup (\eg including challenges and limitations of the synchronization and calibration process between RGB and DVS frames), and (iii) experimenting with different neural network architectures.}


\balance
\bibliography{main}
\bibliographystyle{IEEEtran}

\end{document}